\documentclass[letterpaper]{article} % DO NOT CHANGE THIS
\usepackage{aaai2027}  % DO NOT CHANGE THIS
\usepackage[numbers,sort&compress]{natbib}
\usepackage[hyphens]{url}  % DO NOT CHANGE THIS
\usepackage{graphicx} % DO NOT CHANGE THIS
\usepackage{natbib}  % DO NOT CHANGE THIS AND DO NOT ADD ANY OPTIONS TO IT
\usepackage{caption} % DO NOT CHANGE THIS AND DO NOT ADD ANY OPTIONS TO IT
\usepackage{amssymb}
\usepackage[table]{xcolor}
\definecolor{lightblue}{RGB}{226,239,255}
\usepackage{amsmath}
\usepackage[table]{xcolor}
\usepackage{tabularx}
\newcolumntype{L}{>{\raggedright\arraybackslash}X}

\usepackage{algorithm}
\usepackage{algorithmic}

\nocopyright
\usepackage{newfloat}
\usepackage{listings}
\DeclareCaptionStyle{ruled}{labelfont=normalfont,labelsep=colon,strut=off} % DO NOT CHANGE THIS
\floatstyle{ruled}
\newfloat{listing}{tb}{lst}{}
\floatname{listing}{Listing}
\usepackage{booktabs}

\title{ObjectStream: Latent Objects as Memory Anchors for \\ Streaming Video Understanding}
\author{
    Mingkang Dong\textsuperscript{\rm 1}\equalcontrib,
    Muxin Pu\textsuperscript{\rm 2}\equalcontrib,
    Jie Li\textsuperscript{\rm 3},
    Bohan Guo\textsuperscript{\rm 1},
    Songruo Chen\textsuperscript{\rm 4},
    Bin Ren\textsuperscript{\rm 5},\\
    Xu Zheng\textsuperscript{\rm 6},
    Chen Zhao\textsuperscript{\rm 7},
    Tianwen Qian\textsuperscript{\rm 8},
    Mohamed Elhoseiny\textsuperscript{\rm 7},
    Yuqian Fu\corresponding \textsuperscript{\rm 7}
}

\affiliations{
    \textsuperscript{\rm 1}Universiti Malaya
    \textsuperscript{\rm 2}Monash University
    \textsuperscript{\rm 3}SJTU
    \textsuperscript{\rm 4}Fudan University\\
    \textsuperscript{\rm 5}MBZUAI
    \textsuperscript{\rm 6}Great Bay University
    \textsuperscript{\rm 7}KAUST
    \textsuperscript{\rm 8}ECNU
    
}

\begin{document}
\maketitle
\begin{abstract}
Streaming video understanding requires models to continuously retain useful visual evidence before future questions are known. Existing approaches primarily manage the growing visual context according to token importance, temporal redundancy, or segment-level relevance, but rarely organize evidence around objects that persist and evolve over time. Thus, in this paper, we introduce \textbf{ObjectStream}, a training-free framework that treats \textit{\textbf{latent objects as memory anchors}} for streaming video understanding. ObjectStream induces spatially coherent latent objects directly from frozen Video-LLM representations, links them across frames into persistent anchors, and maintains their histories under a bounded memory budget, without requiring external object detectors or segmentation models. Built on these anchors, ObjectStream preserves three complementary forms of evidence: \textit{persistent object histories}, \textit{transient object changes}, and \textit{recent visual context}. This design enables existing Video Large Language Models (Video-LLMs) to reason over object identities, interactions, and state changes while leaving the underlying model unchanged. Extensive experiments on online streaming and offline long-video benchmarks demonstrate both effectiveness and efficiency. 
In online streaming evaluation, ObjectStream improves Qwen2.5-VL-7B by 10.0 points on OVO-Bench Real-Time Visual Perception, while reducing peak GPU memory and TTFT by approximately 50\%.
On offline long-video benchmarks, it surpasses the full-token baseline while discarding 82.5\% of visual tokens. These results highlight latent objects as a practical and effective organizing principle for compact streaming video memory. \textit{Code is available at https://github.com/DMK041218/ObjectStream.}
\end{abstract}

\begin{figure}[h]
\vspace{0.18in}
\centering
\includegraphics[width=\columnwidth]{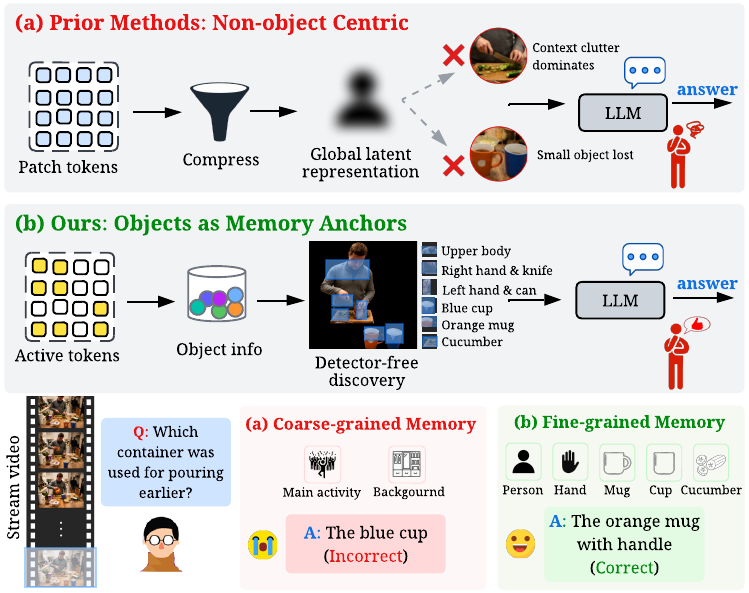}
\caption{
Comparison between existing streaming memory and ObjectStream. Existing methods compress streaming observations into coarse latent representations, whereas ObjectStream organizes memory around latent objects to preserve fine-grained object evidence for long-term reasoning.
}
\vspace{0.05in}
\label{fig:motivation}
\end{figure}
% In online streaming evaluation, ObjectStream improves Qwen2.5-VL-7B by 10.0 points on OVO-Bench Real-Time Visual Perception, while reducing peak GPU memory and TTFT by approximately 50%.


\section{Introduction}
%p1: task intro
Streaming video understanding~\cite{videochatgpt,llavavideo,qwen25vl, wang2025streameqa} requires models to process continuously arriving visual inputs and respond to questions that may be issued at any time. This capability is essential for applications such as live monitoring~\cite{monitoring}, autonomous systems~\cite{vqsop, brodermann2024cafuser, zhao2025survey}, and egocentric vision and wearable assistant~\cite{fu2019embodied, outlook, zhang2026egospotegocentricmultimodalcontrolhandsfree, zhu2026egosound, xu2025tog, zhang2025egonight, li2026egocross, fu2025objectrelator, pan2025v}, and embodied agents and robotic systems~\cite{holoassist,zou2026sis, du2026focusablemonoculardepthestimation, wang2026oflow, wang2026ocra, lin2026la4vla, lin2025evo, wang2026afford, lin2026evo}. Unlike offline video understanding, where the complete video can be revisited after a question is given, streaming models must decide what information to retain before future queries are known, while operating under strict memory and latency constraints.

%p2: discussion of related work 
Recent studies~\cite{savemem,OASIS,HERMES,stc} have extended Video Large Language Models (Video-LLMs) to streaming scenarios from various perspectives, including online adaptation~\cite{longVU,aura}, streaming-oriented inference~\cite{flashvstream,StreamingVLM}, and memory-based context management~\cite{fluxmem,OASIS,HERMES,selectstream,savemem}. Among them, training-free approaches~\cite{fluxmem,streamingTOM,OASIS} are particularly attractive because they can adapt existing offline Video-LLMs without additional training or modification to the underlying model. These methods typically control the growing visual context through frame selection~\cite{simplestream}, visual-token pruning~\cite{fluxmem}, KV-cache compression~\cite{rekv,streamkv}, and temporal or segment-level memory organization~\cite{OASIS,HERMES}. 
By reducing redundant historical information and prioritizing potentially relevant evidence over long video streams, they improve memory efficiency and response latency, facilitating the transition of Video-LLMs from offline video understanding to practical online streaming.

% p3: motivation
However, deciding which visual units or temporal portions to retain addresses only part of the streaming memory problem; how the retained evidence is organized is equally important for subsequent reasoning. While existing designs are effective at preserving coarse-grained contextual and temporal semantics, finer-grained object-level information is often less explicitly organized across time. This limitation is particularly problematic for questions that require tracing an object's identity, interactions, or state changes. These observations motivate an object-centric organization of streaming visual memory.

%p4: core idea of latent objects as memory anchor, & introduce of 3 different kind of memory 
Based on this insight, we introduce \textbf{ObjectStream}, a training-free and plug-and-play visual memory framework that treats \textit{\textbf{latent objects as memory anchors}} for streaming Video-LLMs. Rather than managing historical visual tokens or KV states in isolation, ObjectStream organizes the incoming stream through a \textit{Latent Object-Anchored Memory}. Specifically, it induces spatially coherent object candidates directly from frozen Video-LLM representations, associates them across frames, and maintains their evolving histories under a bounded memory budget. Since these object-like units emerge entirely from the model's latent feature space, without requiring external object detectors or segmentation models, we refer to them as \textit{latent objects}. The resulting cross-frame object states provide a compact semantic index over the stream, enabling the model to preserve object-related evidence beyond locally salient visual content. To complement persistent object histories, ObjectStream introduces two additional components. \textit{Object-Conditioned Temporal Residuals} preserve short-lived but important evidence around abrupt object-level changes, while the \textit{Recent Visual Grounding Window} retains the latest observations for timely visual grounding. Together, these components capture three complementary forms of memory: \textit{persistent object histories}, \textit{transient object changes}, and \textit{recent visual context}, while leaving the underlying Video-LLM unchanged.

%p5: exp highlights
We validate ObjectStream on both online streaming and offline long-video understanding benchmarks. It improves Qwen2.5-VL-7B by 10.0, 4.9, and 2.9 points on OVO-Bench Real-Time Visual Perception, OVO-Bench Backward Tracing, and StreamingBench, respectively. It also surpasses the full-token baseline on offline long-video benchmarks while discarding 82.5\% of visual tokens. These results well demonstrate the effectiveness and token efficiency of ObjectStream.

Our main contributions are summarized as follows:
\begin{itemize}
    \item We propose ObjectStream,  a training-free and plug-and-play visual memory framework for efficient streaming video understanding, without architectural modifications.

    \item We introduce \textit{Latent Object-Anchored Memory}, using latent objects as persistent anchors to organize long-term visual evidence without external detectors or segmenters.

    \item We complement persistent object memory with \textit{Object-Conditioned Temporal Residuals} and a \textit{Recent Visual Grounding Window}, achieving strong performance and token efficiency across online streaming and offline video understanding benchmarks.
\end{itemize}

% Our main contributions are summarized as follows:
% \begin{itemize}
% \item We propose \textbf{ObjectStream}, a training-free and plug-and-play object-centric memory framework that enables existing Video-LLMs to perform efficient streaming video understanding under bounded memory and low-latency constraints, without architectural modifications.

% \item We develop Latent Object-Anchored Memory strategy that converts fragmented patch-level visual tokens into coherent object-level units, providing an explicit semantic structure over incoming video streams without external detectors, segmentation models.

% \item We construct a bounded cross-frame object memory that links object candidates into persistent tracks and maintains them under a fixed budget. By prioritizing objects according to temporal consistency, visual saliency, and recency, ObjectStream preserves informative long-term evidence while preventing unbounded memory growth.

% \item We design Object-Conditioned Temporal Residuals and a recent visual grounding window to capture short-lived event evidence and support low-latency real-time question answering. Experiments on streaming and offline long-video benchmarks show that ObjectStream substantially improves online performance while remaining robust under limited token budgets.
% \end{itemize}

\section{Related Work}

\noindent\textbf{Video Large Language Models.}
Many Video Large Language Models (Video-LLMs) have been proposed to address video-language alignment, temporal modeling, and long-context reasoning. Representative proprietary models include Gemini 1.5 Pro~\cite{gemini} and GPT-4o~\cite{gpt4o}, while widely used open-source models include LLaVA-Video~\cite{llavavideo}, Qwen2-VL~\cite{qwen2vl}, InternVL2~\cite{internvl2}, and LongVU~\cite{longVU}. Despite their strong performance, these models primarily follow an offline inference paradigm, assuming that the complete video is available before question answering. This assumption limits their direct applicability to the more realistic and challenging streaming setting, where visual inputs arrive continuously and future observations are unavailable.

% \noindent\textbf{Streaming Video-LLMs and Visual Memory}
% To enable Video-LLMs in streaming scenarios, existing approaches explore either training-based adaptation or training-free memory management. Training-based methods modify Video-LLMs through additional optimization or streaming-specific architectures. VideoLLM-Online introduces online video instruction tuning to adapt Video-LLMs for continuous visual interaction, while Video Streaming and Flash-VStream design streaming-oriented inference frameworks to improve temporal perception efficiency~\cite{videollmonline,videostreaming,flashvstream}. Although effective, these methods require additional training or architectural modifications.

% Training-free approaches instead preserve the original Video-LLM and focus on managing historical visual information during inference. StreamChat~\cite{streamchat} maintains external visual memories to provide historical context for long-video understanding. ReKV and StreamKV~\cite{rekv,streamkv} optimize KV-cache retention and compression to reduce memory consumption during streaming inference. HERMES, FluxMem, and OASIS~\cite{HERMES,fluxmem,OASIS} further introduce hierarchical memory mechanisms to balance short-term and long-term information retention. Despite their effectiveness, these methods mainly organize historical information around frames, tokens, feature embeddings, or cached states, lacking explicit semantic memory units for preserving persistent object-level continuity.

\noindent\textbf{Streaming Video-LLMs and Visual Memory.}
Existing streaming video understanding models explore either training-based adaptation or training-free memory management. Training-based methods include VideoLLM-Online~\cite{videollmonline}, which introduces online video instruction tuning for continuous video perception and interaction; Dispider~\cite{dispider}, which enables active real-time interaction by disentangling perception, decision, and reaction; and ThinkStream~\cite{thinkstream}, which adopts a Watch--Think--Speak paradigm and learns compressed semantic memory for long-horizon streaming. ViSpeak~\cite{vispeak} and StreamForest~\cite{streamforest} further improve online video understanding through streaming-oriented datasets, training objectives, and model adaptations. These methods generally require additional optimization or model-specific modifications.

Training-free approaches instead preserve the underlying Video-LLM and manage historical visual information directly at inference time. StreamChat~\cite{streamchat} maintains external visual memory to incorporate previous observations, while ReKV and StreamKV~\cite{rekv,streamkv} reduce memory overhead through KV-cache retention and compression. More recent methods explore structured memory management: FluxMem~\cite{fluxmem} introduces hierarchical visual memory to balance long-term history and recent observations; QueryStream~\cite{querystream} performs query-aware token pruning to retain request-relevant evidence; and OASIS~\cite{OASIS} organizes streaming history into event-level structures for on-demand retrieval. However, these methods do not explicitly maintain persistent latent object representations across frames, limiting their ability to preserve object identities and state evolution over long video streams.

% Recent studies extend Video Large Language Models (Video-LLMs) to streaming scenarios through additional training or model adaptation. 

% Although these approaches achieve strong performance in online scenarios, they require additional training and modifications to the original Video-LLMs, limiting their flexibility for directly adapting existing models.

\section{Method}

\begin{figure*}[t]
\centering
\includegraphics[width=1.\textwidth]{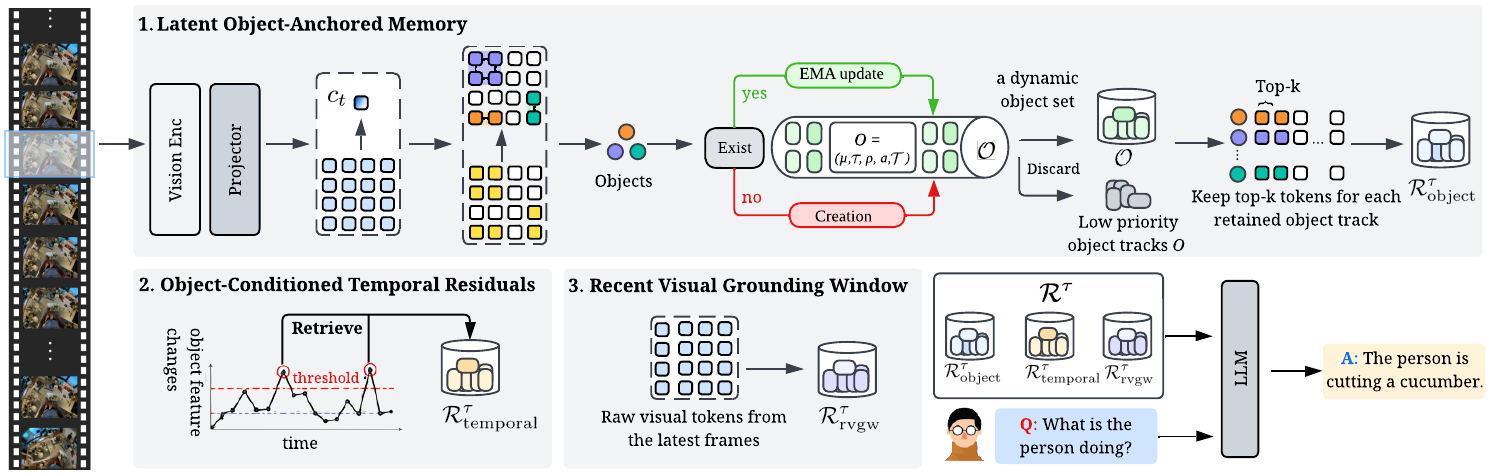}
% \caption{Overview of ObjectStream. ObjectStream first discovers latent object anchors from incoming visual tokens, and then maintains three complementary memory components: Latent Object-Anchored Memory for persistent object histories, Object-Conditioned Temporal Residuals for transient object changes, and a Recent Visual Grounding Window for immediate visual grounding.}
\caption{
Overview of ObjectStream. It comprises three memory modules: \textit{Latent Object-Anchored Memory} discovers and tracks latent object anchors and stores their persistent evidence; \textit{Object-Conditioned Temporal Residuals} preserve transient changes; and the \textit{Recent Visual Grounding Window} retains recent observations for immediate grounding.}
\vspace{0.1in}
\label{fig:method}
\end{figure*}

% \subsection{Overview}
\noindent\textbf{Overview.}
Our design follows three core objectives: maintaining persistent object histories, preserving transient object-level changes, and retaining recent visual context for real-time grounding. The overall framework of our proposed \textbf{ObjectStream} is illustrated in Fig.~\ref{fig:method} and consists of three components. First, \textit{Latent Object-Anchored Memory} discovers spatially coherent latent object candidates from frozen Video-LLM representations, associates them across frames, and maintains their evolving histories under a bounded memory budget. Second, \textit{Object-Conditioned Temporal Residuals} preserve short-lived evidence around abrupt object-level changes. Third, the \textit{Recent Visual Grounding Window} retains the latest observations for timely question answering.

% We propose \textit{ObjectStream}, an object-centric visual memory mechanism for long and streaming video understanding. As illustrated in Fig.~\ref{fig:method}, ObjectStream converts a continuously growing video stream into a compact visual memory that preserves three complementary forms of evidence: persistent object histories, transient object changes, and recent visual context for real-time grounding. Rather than uniformly sampling frames or globally pruning visual tokens, ObjectStream first discovers latent object candidates through spatially constrained clustering, associates them across frames into persistent object anchors, and then maintains three complementary memory components: (1) Latent Object-Anchored Memory (2) Object-Conditioned Temporal Residuals (3) Recent Visual Grounding Window.

Formally, let
\(\mathcal{V}_{\tau}=\{I_t\}_{t=1}^{\tau}\)
denote the video stream observed up to the current timestamp \(\tau\).
ObjectStream processes each frame incrementally upon arrival. At timestamp \(t\leq\tau\), the newly observed frame \(I_t\) is encoded into a set of patch-level visual tokens: $\mathcal{X}_t=\{x_{t,j}\}_{j=1}^{{N}}$, where \(N\) is the number of visual tokens extracted from frame \(I_t\), and each token \(x_{t,j}\in\mathbb{R}^{d}\) is a \(d\)-dimensional visual representation. After processing all observations up to timestamp \(\tau\), ObjectStream maintains the retained visual memory as:
% \quad x_{t,j}\in\mathbb{R}^{d}.$
% \begin{equation}
% \mathcal{X}_t = \{x_{t,j}\}_{j=1}^{\todo{N or N}}
% \quad x_{t,j}\in\mathbb{R}^{d}.
% \label{eq:visual_tokens}
% \end{equation}
% where $t \leq \tau$, $N$ denotes the number of visual tokens in frame $t$, and $d$ is the feature dimension. At timestamp $\tau$, the retained memory is defined as:
\begin{equation}
\mathcal{R}^{\tau}
=
\mathcal{R}^{\tau}_{\mathrm{object}}
\cup
\mathcal{R}^{\tau}_{\mathrm{temporal}}
\cup
\mathcal{R}^{\tau}_{\mathrm{rvgw}},
\label{eq:memory_components}
\end{equation}
% where $\mathcal{R}^{\tau}_{\mathrm{object}}$ stores representative tokens from persistent object tracks, $\mathcal{R}^{\tau}_{\mathrm{temporal}}$ preserves additional raw tokens around abrupt object-level changes, and $\mathcal{R}^{\tau}_{\mathrm{rvgw}}$ denotes the recent visual grounding window that keeps the latest visual observations for immediate question answering.
where $\mathcal{R}^{\tau}_{\mathrm{object}}$, $\mathcal{R}^{\tau}_{\mathrm{temporal}}$, and $\mathcal{R}^{\tau}_{\mathrm{rvgw}}$ are produced by the aforementioned modules, retaining features associated with persistent object histories, transient object-level changes, and recent visual context, respectively.

\subsection{Latent Object-Anchored Memory}

To use latent objects as memory anchors for object-centric evidence retention, this module comprises four stages, denoted as \textit{S1--S4}: \textit{S1: Query-Agnostic Token Saliency Estimation}, \textit{S2: Spatial Clustering for Latent Object Anchor Discovery}, \textit{S3: Cross-Frame Association and Bounded Object-Set Maintenance} and \textit{S4: Object-Centric Evidence Retention}. We describe each stage below.

% \noindent\textit{\textbf{S1: Query-Agnostic Token Saliency Estimation.}}

% \noindent\textit{\textbf{\todo{S2: Spatial Clustering for Latent Object Anchor Discovery.}}}

% \noindent\textit{\textbf{S3: Cross-Frame Association and Bounded Object-Set Maintenance.}}

% \noindent\textit{\textbf{\todo{S4: Object-Centric Evidence Retention.}}}

\begin{figure}[h]
\centering
\includegraphics[width=\columnwidth]{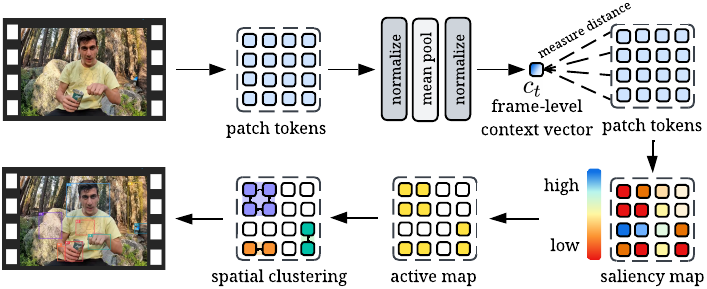}
\caption{Illustration of latent object discovery. ObjectStream discovers latent object candidates directly from frozen visual tokens by estimating query-agnostic saliency and spatially clustering salient regions. The resulting latent objects provide the semantic object anchors for memory construction. 
}
\label{fig:discovery}
\vspace{0.1in}
\end{figure}
% \subsection{Latent Anchor Discovery and Tracking}
\label{sec:object_discovery}
% To establish latent objects as memory anchors without relying on external object detectors or segmentation models, ObjectStream first extracts spatially coherent object candidates from individual frames and then tracks them across time into persistent object anchors.

% \paragraph{Query-agnostic token saliency.}
\noindent\textit{\textbf{S1: Query-Agnostic Token Saliency Estimation.}}
For each frame $I_t$, we first summarize its overall visual content using a
frame-level context vector $c_t$. Specifically, we $\ell_2$-normalize each visual
token and aggregate the normalized tokens into a global context representation : 
\begin{equation}
\begin{aligned}
\hat{x}_{t,j} &= \frac{x_{t,j}}{\|x_{t,j}\|_2}, \\
c_t &= \frac{\sum_{j=1}^{N}\hat{x}_{t,j}}
{\left\|\sum_{j=1}^{N}\hat{x}_{t,j}\right\|_2}.
\end{aligned}
\label{eq:frame_context}
\end{equation}

The context vector $c_t$ captures the dominant visual semantics of frame
$I_t$. We then estimate the saliency of each token by its cosine
dissimilarity from this frame-level context:
\begin{equation}
s_{t,j}
=
1-\mathrm{cos}(\hat{x}_{t,j},c_t)
=
1-\hat{x}_{t,j}^{\top}c_t .
\label{eq:saliency}
\end{equation}

The score $s_{t,j}$ measures how visually distinctive token $x_{t,j}$ is within the current frame. Tokens that deviate more from the global frame representation are more likely to correspond to informative objects, persons, or regions. We then select active token candidates using a per-frame quantile threshold:
\begin{equation}
\mathcal{A}_t = \left\{ j \mid s_{t,j} \geq \mathrm{Quantile}_{q}\left(\mathcal{S}_t\right) \right\},
\label{eq:active_tokens}
\end{equation}
where $\mathcal{S}_t = \{s_t,_j\}_{j=1}^{{N}}$ is the set of saliency scores for all visual tokens in frame $t$, and $\mathrm{Quantile}_q(\mathcal{S}_t)$ denotes the $q$-th quantile of
saliency scores within frame $t$; thus larger $q$ selects fewer, more salient tokens as active object candidates.
% For each frame, we compute a frame-level context vector $c_t$ from normalized visual token features and define the saliency of each token as its cosine distance from this context:
% \begin{equation}
% \begin{aligned}
% \hat{x}_{t,j} &= \mathrm{normalize}(x_{t,j}), \\
% c_t &= \mathrm{normalize}\left(\sum_{j=1}^{N}\hat{x}_{t,j}\right), \\
% s_{t,j} &= 1-\hat{x}_{t,j}^{\top}c_t .
% \end{aligned}
% \label{eq:saliency}
% \end{equation}
% The score $s_{t,j}$ measures how visually distinctive token $x_{t,j}$ is within the current frame. Tokens that deviate more from the global frame representation are more likely to correspond to informative objects, persons, or regions. We then select active token candidates using a per-frame quantile threshold:
% \begin{equation}
% \mathcal{A}_t = \left\{ j \mid s_{t,j} \geq \mathrm{Quantile}_{q}\left(\mathcal{S}_t\right) \right\}.
% \label{eq:active_tokens}
% \end{equation}
% where $\mathcal{S}_t = \{s_t,j\}_{j=1}^{N}$ is the set of saliency scores for all visual tokens in frame $t$, and $q$ controls the selectivity of object candidate extraction.

% \paragraph{Spatial clustering into object candidates.}
\noindent\textit{\textbf{S2: Spatial Clustering for Latent Object Anchor Discovery.}}
As shown in Fig.~\ref{fig:discovery}, the active tokens indexed by $\mathcal{A}_t$ are still patch-level candidates and may be spatially fragmented. To form coherent object-level regions, ObjectStream exploits the 2D patch-grid structure of visual tokens. Let $p_{t,j}$ denote the spatial position of token $x_{t,j}$. We build an undirected graph $\mathcal{G}_t$ over active tokens, where edges connect spatially adjacent tokens:
\begin{equation}
\begin{aligned}
\mathcal{G}_t &= (\mathcal{A}_t,\mathcal{E}_t), \\
(j,k)\in\mathcal{E}_t &\Longleftrightarrow j,k\in\mathcal{A}_t  \land  p_{t,k}\in\mathcal{N}(p_{t,j}), \\
\mathcal{C}_t &= \mathrm{CC}(\mathcal{G}_t).
\end{aligned}
\label{eq:spatial_clustering}
\end{equation}

Here, $\mathcal{E}_t$ is the edge set,
$\mathcal{N}_4(p_{t,j})$ denotes the 4-connected spatial neighborhood
of patch $p_{t,j}$ on the patch grid, including its upper, lower, left,
and right neighbors. $\mathrm{CC}(\cdot)$ denotes connected components,
and $\mathcal{C}_t=\{C_{t,m}\}_{m=1}^{M_t}$ is the set of frame-level
object candidates.
\begin{equation}
\mu = f(C_{t,m}) = \frac{\sum_{j \in C_{t,m}} s_{t,j} x_{t,j}}{\left\| \sum_{j \in C_{t,m}} s_{t,j} x_{t,j} \right\|_2}.
\label{eq:cluster_feature}
\end{equation}

This transforms scattered salient patch tokens into coarse object-level units without requiring external detectors or segmentation annotations.

% \paragraph{Cross-frame tracking and bounded object set.}
\noindent\textit{\textbf{S3: Cross-Frame Association and Bounded Object-Set Maintenance.}}
After obtaining frame-level object candidates, ObjectStream associates them across frames to build persistent object tracks. We maintain a dynamic object set $\mathcal{O}=\{O_i\}$, where each object track $O_i$ stores an object feature $\mu_i$, the last observed timestamp $\tau_i$, a temporal consistency score $\rho_i$, an object saliency score $a_i$, and a historical token set $\mathcal{T}_i$. The temporal consistency score $\rho_i$ is computed as the cosine similarity between the track feature and the candidate feature at its most recent match, indicating the confidence of cross-frame association. Upon each successful match, the temporal consistency score $\rho_{i}$ is updated as the cosine similarity. The object saliency score $a_i$ is inherited from the token-level saliency introduced in S2 by running averaging the saliency values of the visual tokens belonging to the corresponding object candidate. 

For each frame-level candidate $C_{t,m}$, ObjectStream first finds the most similar existing object track:
\begin{equation}
i^{\star} = \arg\max_i \cos\left(f(C_{t,m}),\mu_i\right).
\label{eq:object_match_index}
\end{equation}

If $\cos(f(C_{t,m}),\mu^{t-1}_{i^{\star}})\geq\theta$, the candidate is assigned to $O_{i^{\star}}$ and the object feature is updated by Exponential Moving Average (EMA):
\begin{equation}
\mu_{i^*}^{t} = \frac{\beta \mu_{i^*}^{t-1} + (1 - \beta) f(C_{t,m})}{\left\| \beta \mu_{i^*}^{t-1} + (1 - \beta) f(C_{t,m}) \right\|_2}.
\label{eq:object_feature_update}
\end{equation}

If no existing object satisfies the matching threshold, ObjectStream initializes a new object track. For a matched object, we set $\tau_{i^*}$ to the current timestamp, update its temporal consistency $\rho_{i^*}$ and average saliency $a_{i^*}$, and insert all tokens in the matched component into $\mathcal{T}_{i^*}$.

Since streaming videos continuously introduce new objects, the object set must remain bounded. When the number of maintained tracks exceeds the budget $M$, ObjectStream evicts low-priority objects according to recency, saliency, and temporal consistency.
\begin{equation}
\begin{aligned} 
\mathrm{rank}(O_i) &= (\tau_i, \ a_i, \rho_i), \\
\mathcal{O} &= \mathrm{Top\text{-}M}_{O_i\in\mathcal{O}}\left(\mathrm{rank}(O_i)\right).
\end{aligned}
\label{eq:object_eviction}
\end{equation}

The ranking is applied in lexicographic order, prioritizing object anchors that are recently observed, visually salient, and temporally consistent.

% \subsection{Latent Object-Anchored Memory}
\noindent\textit{\textbf{S4: Object-Centric Evidence Retention.}}
Once persistent object tracks in the set $\mathcal{O}$ are maintained under a fixed budget, ObjectStream constructs the long-term object-centric evidence $\mathcal{R}^{\tau}_{\mathrm{object}}$ by selecting representative tokens from each retained object anchor.

Object eviction and token retention operate at different granularities: the bounded object set $\mathcal{O}$ determines which object tracks are maintained up to timestamp $\tau$, while token retention is performed over the historical token pool $\mathcal{T}_i^{\tau}$ of each retained track $O_i \in \mathcal{O}$. For each retained object $O_i$, its historical token pool contains token indices $(t,j)$ assigned to this object up to timestamp $\tau$. Instead of globally selecting top-scoring tokens from the whole video, which may be dominated by a few salient regions or frames, ObjectStream allocates token retention capacity at the object level:
\begin{equation}
\begin{aligned}
\mathcal{I}^{\tau}_i &= \operatorname{Top\text{-}K}_{(t,j)\in\mathcal{T}^{\tau}_i,\; t\leq\tau}\left(s_{t,j}, k_i\right), \\
\mathcal{R}^{\tau}_{\mathrm{object}} &= \bigcup_{O_i\in\mathcal{O}^{}} \left\{ x_{t,j} \mid (t,j)\in\mathcal{I}^{\tau}_i \right\}.
\end{aligned}
\label{eq:object_memory}
\end{equation}

The operator $\operatorname{Top\text{-}K}$ ranks tokens within the object-specific pool $\mathcal{T}_i^{\tau}$ by their saliency scores $s_{t,j}$ and returns the top-$k_i$ token indices. The retained object memory $\mathcal{R}^{\tau}_{\mathrm{object}}$ then collects the corresponding visual tokens from all maintained object tracks, forming a compact yet comprehensive representation of persistent visual histories.

\subsection{Object-Conditioned Temporal Residuals}

Latent Object-Anchored memory preserves persistent visual semantics by smoothing object representations over time. However, this smoothing may suppress short-lived but important changes, such as motion, action transitions, object appearance or disappearance, state changes, and interaction changes. 
Thus, this module is further proposed to detect abrupt object-level feature changes and preserve additional raw tokens around the corresponding timestamps.

For each object track $O_i$, we denote its matched sequence of historical observations as $\{z_{i,l}\}_{l=1}^{L_i}$. Specifically, if the frame-level object candidate $f(C_{t,m})$ is successfully assigned to track $O_i$ at a timestamp $t$, this observation is mapped as $z_{i,l} = f(C_{t,m})$, where $l$ indexes the chronologically ordered observation sequence up to length $L_i$, where $2 \le l \le L_i$, and $t_{i,l} = t$ preserves its corresponding timestamp. We then compute the feature change between neighboring observations and use an object-specific adaptive threshold to detect temporal events:
\begin{equation}
\begin{aligned}
\Delta_{i,l}
&=
1-\cos(z_{i,l},z_{i,l-1}), \\
\gamma_{i, l}
&=
\mathrm{mean}
\left(
\{\Delta_{i,k}\}_{k=2}^{l}
\right)
+
\mathrm{std}
\left(
\{\Delta_{i,k}\}_{k=2}^{l}
\right), \\
\mathcal{B}^{\tau}
&=
\left\{
(i,t_{i,l})
\mid
\Delta_{i,l}>\gamma_i,\;
t_{i,l}\leq\tau
\right\}.
\end{aligned}
\label{eq:temporal_detection}
\end{equation}

Here, $\Delta_{i,l}$ measures the temporal change of object $O_i$ between two consecutive matched observations, $\gamma_{i,l}$ is its adaptive temporal threshold, and $\mathcal{B}^{\tau}$ denotes the temporal events detected up to timestamp $\tau$.

For each temporal event $(i,t)\in\mathcal{B}^{\tau}$, ObjectStream retrieves the original visual tokens assigned to the same object within a local temporal window
$\mathcal{W}(t,r)=\{u\mid t-r\leq u\leq t+r,\; u\leq\tau\}$, where $r$ denotes the temporal window frame offset that defines the local neighborhood around event timestamp $t$.
Within this object-specific temporal neighborhood, we select token indices according to their saliency scores:
\begin{equation}
\begin{aligned}
\mathcal{I}^{\tau}_{i,t}
&=
\operatorname{TopK}_{(u,j)\in\mathcal{T}^{\tau}_i,\;u\in\mathcal{W}(t,r)}
\left(s_{u,j}, k_b\right), \\
\mathcal{R}^{\tau}_{\mathrm{temporal}}
&=
\bigcup_{(i,t)\in\mathcal{B}^{\tau}}
\left\{
x_{u,j}
\mid
(u,j)\in\mathcal{I}^{\tau}_{i,t}
\right\}.
\end{aligned}
\label{eq:temporal_memory}
\end{equation}

In Eq.\ref{eq:temporal_memory} $\operatorname{Top\text{-}K}$ returns the indices of the top-$k_b$ tokens ranked by saliency within the local temporal neighborhood, and $\mathcal{R}^{\tau}_{\mathrm{temporal}}$ stores the corresponding raw visual tokens. 

Unlike object-level representative tokens, temporal residual tokens preserve raw visual evidence around rapid object-level changes. Therefore, they complement the smooth long-term object memory and improve the ability to reason about short-term temporal events.

\begin{table*}[t]
\centering
\small
\setlength{\tabcolsep}{1.4pt}
\renewcommand{\arraystretch}{0.95}
\begin{tabular}{@{}l c|rrrrrrr|r|rrrrrrrrrrr@{}}
\toprule
\textbf{Method} & \textbf{Frames}
& \multicolumn{7}{c|}{\textbf{OVO-Bench Real-Time}}
& \textbf{BT}
& \multicolumn{11}{c}{\textbf{StreamingBench}} \\
\cmidrule(lr){3-9} \cmidrule(lr){10-10} \cmidrule(lr){11-21}
&
& \textbf{OCR} & \textbf{ACR} & \textbf{ATR} & \textbf{STU} & \textbf{FPD} & \textbf{OJR} & \textbf{Avg.}
& \textbf{Avg.}
& \textbf{OP} & \textbf{CR} & \textbf{CS} & \textbf{ATP} & \textbf{EU}
& \textbf{TR} & \textbf{PR} & \textbf{SU} & \textbf{ACP} & \textbf{CT} & \textbf{Avg.} \\
\midrule
\multicolumn{21}{l}{\textit{Proprietary Models}} \\
\midrule
Gemini 1.5 Pro & 1 fps
& 85.9 & 67.0 & 79.3 & 58.4 & 63.4 & 62.0 & 69.3
& 62.5
& 79.0 & 80.5 & 83.5 & 79.7 & 80.0 & 84.7 & 77.8 & 64.2 & 72.0 & 48.7 & 75.7 \\
GPT-4o & 64
& 69.8 & 64.2 & 71.6 & 51.1 & 70.3 & 59.8 & 64.5
& 60.8
& 77.1 & 80.5 & 83.9 & 76.5 & 70.2 & 83.8 & 66.7 & 62.2 & 69.1 & 49.2 & 73.3 \\

\midrule
\multicolumn{21}{l}{\textit{Open-source Offline MLLMs}} \\
\midrule
LLaVA-Video & 64
& 69.8 & 59.6 & 66.4 & 50.6 & 72.3 & 61.4 & 63.3
& 41.7
& -- & -- & -- & -- & -- & -- & -- & -- & -- & -- & -- \\
Qwen2-VL & 64
& 69.1 & 53.2 & 63.8 & 50.6 & 66.3 & 60.9 & 60.7
& 48.6
& -- & -- & -- & -- & -- & -- & -- & -- & -- & -- & -- \\
InternVL2 & 64
& 68.5 & 58.7 & 69.0 & 44.9 & 67.3 & 56.0 & 60.7
& 44.0
& 68.1 & 60.9 & 69.4 & 77.1 & 67.7 & 62.9 & 59.3 & 53.3 & 55.0 & \underline{56.5} & 63.7 \\
LongVU & 1 fps
& 55.7 & 49.5 & 59.5 & 48.3 & 68.3 & 63.0 & 57.4
& 39.5
& -- & -- & -- & -- & -- & -- & -- & -- & -- & -- & -- \\

\midrule
\multicolumn{21}{l}{\textit{Open-source Online MLLMs (Training-Based)}} \\
\midrule
VideoLLM-Online-8B & 2 fps
& 8.1 & 23.9 & 12.1 & 14.0 & 45.5 & 21.2 & 20.8
& 17.7
& 39.1 & 40.1 & 34.5 & 31.1 & 46.0 & 32.4 & 31.5 & 34.2 & 42.5 & 27.9 & 36.0 \\
Dispider-7B & 1 fps
& 57.7 & 49.5 & 62.1 & 44.9 & 61.4 & 51.6 & 54.6
& 36.1
& 74.9 & 75.5 & 74.1 & 73.1 & 74.4 & 59.9 & 76.1 & 62.9 & 62.2 & 45.8 & 67.6 \\
Flash-VStream-7B & 1 fps
& 25.5 & 32.1 & 29.3 & 33.7 & 29.7 & 28.8 & 29.9
& 25.4
& 25.9 & 43.6 & 24.9 & 23.9 & 27.3 & 13.1 & 18.5 & 25.2 & 23.9 & 48.7 & 23.2 \\

ViSpeak & 1 fps
& 75.2 & 58.7 & 71.6 & 51.1 & 74.3 & \underline{66.9} & 66.3
& \textbf{57.5}
& 79.8 & \textbf{88.3} & \underline{83.3} & 81.1 & 76.4 & 75.1 & 70.4 & 65.9 & \textbf{77.3} & 34.2 & 74.4 \\
ThinkStream & 1 fps
& 85.2 & 64.2 & 69.8 & 49.4 & 69.3 & 64.1 & 67.0
& 52.3
& -- & -- & -- & -- & -- & -- & -- & -- & -- & -- & -- \\
TimeChat-Online-7B & 1 fps
& 75.2 & 46.8 & 70.7 & 47.8 & 69.3 & 61.4 & 61.9
& 41.7
& 80.8 & 79.7 & 80.8 & 83.3 & 74.8 & 78.8 & 78.7 & 64.2 & 68.8 & \textbf{58.0} & 75.3 \\
StreamForest-7B & 1 fps
& 68.5 & 53.2 & 71.6 & 47.8 & 65.4 & 60.9 & 61.2
& 52.0
& \textbf{83.1} & 82.8 & 82.7 & \underline{84.3} & \underline{77.5} & 78.2 & 76.9 & 69.1 & \underline{75.6} & \underline{54.4} & \textbf{77.3} \\

\midrule
\multicolumn{21}{l}{\textit{Open-source Online MLLMs (Training-Free)}} \\
\midrule
Qwen2.5-VL-3B$^\dagger$ & 1 fps
& 76.5 & 44.0 & 67.2 & 42.1 & 68.3 & 62.0 & 60.0
& 42.0
& 76.2 & 68.8 & 75.4 & 79.2 & 73.0 & 72.3 & 71.3 & 61.4 & 71.6 & 26.1 & 68.0 \\
\quad + FluxMem & 1 fps
& 83.2 & 56.9 & 67.2 & 47.8 & 68.3 & 63.6 & 64.5
& 42.1
& 72.5 & 73.8 & 73.8 & 79.4 & 72.6 & 79.7 & 76.9 & 64.2 & 65.1 & 42.6 & 70.9 \\
\quad + \textbf{ObjectStream (Ours)} & 1 fps
& \underline{87.3} & 68.8 & \underline{73.3} & \underline{54.5} & 66.3 & 65.2 & \underline{68.6}
& 42.5
& 72.7 & 69.8 & 79.8 & 80.4 & 73.9 & 82.8 & 65.7 & \underline{71.1} & 72.0 & 34.6 & 72.4 \\

\midrule
Qwen2.5-VL-7B$^\dagger$ & 1 fps
& 79.2 & 53.2 & 67.2 & 51.7 & 71.3 & 57.1 & 63.3
& 44.6
& 78.3 & 80.5 & 79.8 & 82.4 & 75.5 & 80.4 & 74.1 & 62.6 & 67.6 & 51.1 & 73.9 \\
\quad + FluxMem & 1 fps
& 81.2 & 59.6 & 70.7 & 53.4 & \underline{75.2} & 63.0 & 67.2
& 46.8
& 80.2 & 81.1 & 81.4 & \textbf{85.3} & \textbf{78.0} & \underline{83.8} & \underline{80.6} & 65.9 & 69.6 & 52.1 & 76.4 \\

\quad + QueryStream & 1 fps
& 75.2 & 49.5 & 69.8 & 50.0 & 71.3 & 62.5 & 63.1 & 44.9
& \underline{82.4} & \underline{84.4} & 79.2 & 82.4 & \textbf{78.0} & 81.3 & 78.7 & 65.0 & 69.3 & 47.3 & 75.3 \\

\quad + OASIS & --
& 85.2 & \textbf{72.5} & 66.4 & 52.3 & 67.3 & 64.7 & 64.7 & \underline{52.6} &
-- & -- & -- & -- & -- & -- & -- & -- & -- & -- & 70.6 \\

\quad + \textbf{ObjectStream (Ours)}  & 1 fps
& \textbf{89.9} & \underline{70.6} & \textbf{74.1} & \textbf{59.0} & \textbf{75.3} & \textbf{70.7} & \textbf{73.3}
& 49.5
& 74.1 & 80.2 & \textbf{88.6} & 82.6 & 77.1 & \textbf{85.0} & \textbf{82.4} & \textbf{73.6} & 73.2 & 44.2 & \underline{76.8} \\
\bottomrule
\end{tabular}
\caption{Performance comparison on OVO-Bench and StreamingBench. For OVO-Bench, we report real-time visual perception scores and the average backward tracing score. For StreamingBench, we report real-time category scores and the average score. \textbf{Bold} indicates the best result, and \underline{underlining} indicates the second-best result. $\dagger$ indicates the backbone model.}
\label{tab:online_benchmarks}
\vspace{0.1in}
\end{table*}

\begin{table}[!h]
\centering
\small
\setlength{\tabcolsep}{2.4pt}
\renewcommand{\arraystretch}{1.10}

\begin{tabular}{ccc|cccc}
\toprule

\multicolumn{3}{c|}{\textbf{Component}}
& \textbf{VME}
& \textbf{Ego}
& \textbf{OVO}
& \textbf{Streaming}
\\

\textbf{RVGW}
& \textbf{LOAM}
& \textbf{OCTR}
& \textbf{Long}
& \textbf{Schema}
& \textbf{Bench}
& \textbf{Bench}
\\

\midrule
\checkmark &            &            & 52.3 & 58.6 & 62.3 & 73.9 \\
           & \checkmark &            & 51.1 & 58.3 & 62.0 & 74.8 \\
\checkmark & \checkmark &            & 53.4 & 59.4 & 60.6 & 76.5 \\
           & \checkmark & \checkmark & 53.8 & 60.4 & 62.1 & 74.9 \\
\checkmark & \checkmark & \checkmark
& \textbf{54.0}
& \textbf{60.8}
& \textbf{65.4}
& \textbf{76.8}\\
\bottomrule
\end{tabular}
\caption{Component ablation of ObjectStream.}
\label{tab:component_ablation}
\end{table}

\begin{table}[!h]
\centering
\small
\setlength{\tabcolsep}{1.9pt}
\renewcommand{\arraystretch}{0.95}
\begin{tabular}{@{}lccc@{}}
\toprule
\textbf{Model} & \textbf{\#Frames} & \textbf{VideoMME-L} & \textbf{EgoSchema} \\
\midrule
\multicolumn{4}{c}{\textit{\textbf{Proprietary MLLMs}}} \\
\midrule
Gemini 1.5 Pro & 1 fps & 62.5 & 69.3 \\
GPT-4o & 64 & 60.7 & 64.5 \\
\midrule
\multicolumn{4}{c}{\textit{\textbf{Open-source Offline MLLMs}}} \\
\midrule
LLaVA-Video-7B & 32 & -- & 57.3 \\
Qwen2-VL-7B & 64 & -- & \textbf{66.7} \\
LongVU-7B & 1 fps & \textbf{60.6} & 58.2 \\
\midrule
\multicolumn{4}{c}{\textit{\textbf{Open-source Online MLLMs}}} \\
\midrule
TimeChat-Online-7B & 1 fps & 48.4 & \underline{61.9} \\
Dispider-7B & 1 fps & 49.7 & 55.6 \\
Vista & 1 fps & -- & 58.7 \\
\midrule
\multicolumn{4}{c}{\textit{\textbf{Training-free Offline-to-Online Methods}}} \\
\midrule
Qwen2.5-VL-3B$^\dagger$ & 1 fps & 49.3 & 54.9 \\
\quad + FluxMem & 1 fps & 51.2 & 57.8 \\
\quad + \textbf{ObjectStream (Ours)}  & 1 fps & 52.9 & 57.7 \\
\midrule
Qwen2.5-VL-7B$^\dagger$ & 1 fps & 51.3 & 58.5 \\
\quad + FluxMem & 1 fps & 53.9 & 60.1 \\
\quad + QueryStream & 1 fps & 52.9 & -- \\
\quad + \textbf{ObjectStream (Ours)}  & 1 fps & \underline{54.0} & 60.8 \\
\bottomrule
\end{tabular}
\caption{Performance comparison (\%) on offline benchmarks.}
\vspace{0.1in}
\label{tab:offline_benchmarks}
\end{table}

% \subsection{Recent Visual Grounding Window and Answer Generation}
\subsection{Recent Visual Grounding Window}
Although object memory and temporal residuals preserve long-term semantics and short-term dynamics, streaming questions may also require the latest visual state. For example, questions about what is currently happening or which object is now visible depend strongly on the most recent frames. To support such real-time questions, ObjectStream maintains a lightweight \textit{Recent Visual Grounding Window} (rvgw), which directly preserves raw visual tokens from the latest $L$ frames:
\begin{equation}
\mathcal{R}_{\mathrm{rvgw}}^{\tau}
=
\bigcup_{t=T-L+1}^{T}
\mathcal{X}_t.
\label{eq:rvgw_memory}
\end{equation}

The rvgw complements the compressed object memory by providing high-fidelity local evidence for the current visual state, while the object and temporal residual memories preserve compact historical context.

\noindent\textit{\textbf{Inference.}} Given a textual question $Q$, the language model generates the answer by conditioning on the retained visual memory:
\begin{equation}
\hat{Y}
=
\mathrm{LLM}(Q,\mathcal{R}^{\tau}).
\label{eq:llm_generation}
\end{equation}

% By combining spatially constrained object clustering, bounded cross-frame object memory, object-aware token retention, Object-Conditioned Temporal Residuals, and a recent visual grounding window, ObjectStream provides a compact yet expressive visual representation for long and streaming video understanding.

By integrating persistent object histories, transient object-level changes, and recent visual context, ObjectStream provides the underlying Video-LLM with a compact yet expressive representation of the observed stream. This unified memory supports both historical reasoning and grounding in the current visual state under a bounded visual-token budget.

\section{Experiments}
\noindent\textbf{Implementation Details.}
For all experiments, we build our method on Qwen2.5-VL-3B and 7B~\cite{qwen25vl}. For online benchmarks, we sample videos at 1 fps and set the maximum video length to 256 frames. We use 8 frames as \textit{the Recent Visual Grounding Window} (rvgw) for real-time question answering and compress older frames into object-centric memory. We keep up to 64 objects with 8 tokens per object, and further enable temporal residuals with at most 32 events, radius 1, and extra 2 tokens per selected frame.
For offline benchmarks, we limit the maximum sequence length to 1024 frames. We use a larger object memory budget to preserve longer-range temporal context, keeping up to 512 objects with 80 tokens per object. We preserve the most recent 80 frames as the rvgw, while compressing older visual content into object-centric long-term memory. %For temporal residuals, we keep up to 328 temporal events, with at most 24 events per object, a temporal radius of 1, and 32 temporal tokens per selected frame. We set default saliency quantile $\rho$ to 0.6, matching threshold $\theta$to 0.72, EMA Decay $\beta$ to 0.90. All experiments are conducted on 8 NVIDIA A6000 GPUs

\noindent\textbf{Benchmarks.}
We evaluate our method on both online and offline benchmarks.
For online evaluation, we use StreamingBench~\cite{streamingbench} and OVO-Bench~\cite{ovobench}. StreamingBench evaluates streaming video question answering under causal access, requiring the model to answer questions incrementally as video frames arrive over time. OVO-Bench focuses on online video understanding with an emphasis on real-time reasoning, covering abilities such as object perception, temporal reasoning, causal understanding, and event-based decision making under constrained memory.
For offline evaluation, we use EgoSchema~\cite{egoschema} and VideoMME-Long ~\cite{videomme}. These benchmarks evaluate holistic video understanding when the entire video is available, with a particular focus on long-range temporal reasoning, egocentric understanding, and comprehensive multi-step video QA.

\noindent\textbf{Results on Online Benchmark.}
We compare ObjectStream with strong proprietary models, open-source offline Video-LLMs, training-based online Video-LLMs, and training-free online methods, including Gemini 1.5 Pro~\cite{gemini}, GPT-4o~\cite{gpt4o}, LLaVA-Video~\cite{llavavideo}, Qwen2-VL~\cite{qwen2vl}, InternVL2~\cite{internvl2}, LongVU~\cite{longVU}, VideoLLM-Online~\cite{videollmonline}, Dispider~\cite{dispider}, Flash-VStream~\cite{flashvstream}, ViSpeak~\cite{vispeak}, ThinkStream~\cite{thinkstream}, TimeChat-Online~\cite{timechatonline}, StreamForest~\cite{streamforest}, QueryStream~\cite{querystream}, OASIS~\cite{OASIS} and FluxMem~\cite{fluxmem}.

\noindent\textbf{Main Results on Streaming Benchmarks.}
As shown in Tab.~\ref{tab:online_benchmarks}, ObjectStream consistently improves streaming video understanding across online benchmarks without fine-tuning or architectural modification, demonstrating its effectiveness as a lightweight and plug-and-play visual memory mechanism for Video-LLMs.

On OVO-Bench, ObjectStream achieves substantial improvements over Qwen2.5-VL baselines. With the 7B backbone, it improves the real-time visual perception score from 63.3 to 73.3 (+10.0) and the Backward Tracing score from 44.6 to 49.5, indicating better preservation of historical visual cues. Compared with FluxMem, ObjectStream further improves the overall score from 60.4 to 65.4 on the 7B backbone and from 54.0 to 60.2 on the 3B backbone;
On StreamingBench, ObjectStream improves the average score from 73.9 to 76.8 with the 7B backbone and from 68.0 to 72.4 with the 3B backbone. These results demonstrate that latent object memory construction provides a general and effective solution for online perception and temporal reasoning.

\paragraph{Effect of Memory Components.}
To evaluate the contributions of Recent Visual Grounding Window (RVGW), Latent Object-Anchored Memory (LOAM), and Object-Conditioned Temporal Residuals (OCTR), we progressively enable each module. As shown in Tab.~\ref{tab:component_ablation}, RVGW mainly improves StreamingBench by preserving recent visual observations, while LOAM enhances OVO-Bench through structured object memory. Combining the two consistently improves performance across both streaming and offline settings. Adding OCTR further preserves abrupt object changes, resulting in the best overall performance.

\noindent\textbf{Extension to Offline Long-video Benchmarks.} We also evaluate ObjectStream on offline long-video benchmarks to verify whether the proposed memory design remains effective in the offline scenario. As shown in Tab.~\ref{tab:offline_benchmarks}, ObjectStream consistently improves the original Qwen2.5-VL baselines on both VideoMME-Long and EgoSchema. With Qwen2.5-VL-7B, our method improves VideoMME-Long from 51.3 to 54.0 and EgoSchema from 58.5 to 60.8, achieving gains of 2.7 and 2.3 points, respectively. With Qwen2.5-VL-3B, ObjectStream also improves VideoMME-Long from 49.3 to 52.9 and EgoSchema from 54.9 to 57.7, showing that the benefit is consistent across model scales. These results demonstrate that ObjectStream is not limited to online streaming inference. %By organizing visual evidence around persistent objects and preserving temporal cues, our method also provides an effective memory mechanism for offline long-video understanding.

% \subsection{Ablation Studies}

% \paragraph{Effect of Memory Components.}
% \begin{table}[h]
% \centering
% \caption{Component ablation of ObjectStream.}
% \label{tab:component_ablation}
% \small
% \setlength{\tabcolsep}{2.4pt}
% \renewcommand{\arraystretch}{1.10}
% \begin{tabular}{ccc|cccc}
% \toprule
% \multicolumn{3}{c|}{\textbf{Component}}
% & \textbf{VME}
% & \textbf{Ego}
% & \textbf{OVO}
% & \textbf{Streaming}
% & \textbf{RVGW}
% & \textbf{LOAM}
% & \textbf{OCTR}
% & \textbf{Long}
% & \textbf{Schema}
% & \textbf{Bench}
% & \textbf{Bench}\\
% \midrule
% \checkmark &            &            & 52.3 & 58.6 & 62.3 & 73.9 \\
%            & \checkmark &            & 51.1 & 58.3 & 62.0 & 74.8 \\
% \checkmark & \checkmark &            & 53.4 & 59.4 & 60.6 & 76.5 \\
%            & \checkmark & \checkmark & 53.8 & 60.4 & 62.1 & 74.9 \\
% \checkmark & \checkmark & \checkmark
% & \textbf{54.0}
% & \textbf{60.8}
% & \textbf{65.4}
% & \textbf{76.8}\\
% \bottomrule
% \end{tabular}
% \end{table}

% To evaluate the contributions of Recent Visual Grounding Window (RVGW), Latent Object-Anchored Memory (LOAM), and Object-Conditioned Temporal Residuals (OCTR), we progressively enable each module. As shown in Tab.~\ref{tab:component_ablation}, RVGW mainly improves StreamingBench by preserving recent visual observations, while LOAM enhances OVO-Bench through structured object memory. Combining the two consistently improves performance across both streaming and offline settings. Adding OCTR further preserves abrupt object changes, resulting in the best overall performance.

\begin{figure*}[!t]
    \centering
    \vspace{0.1in}
    \includegraphics[width=\textwidth]{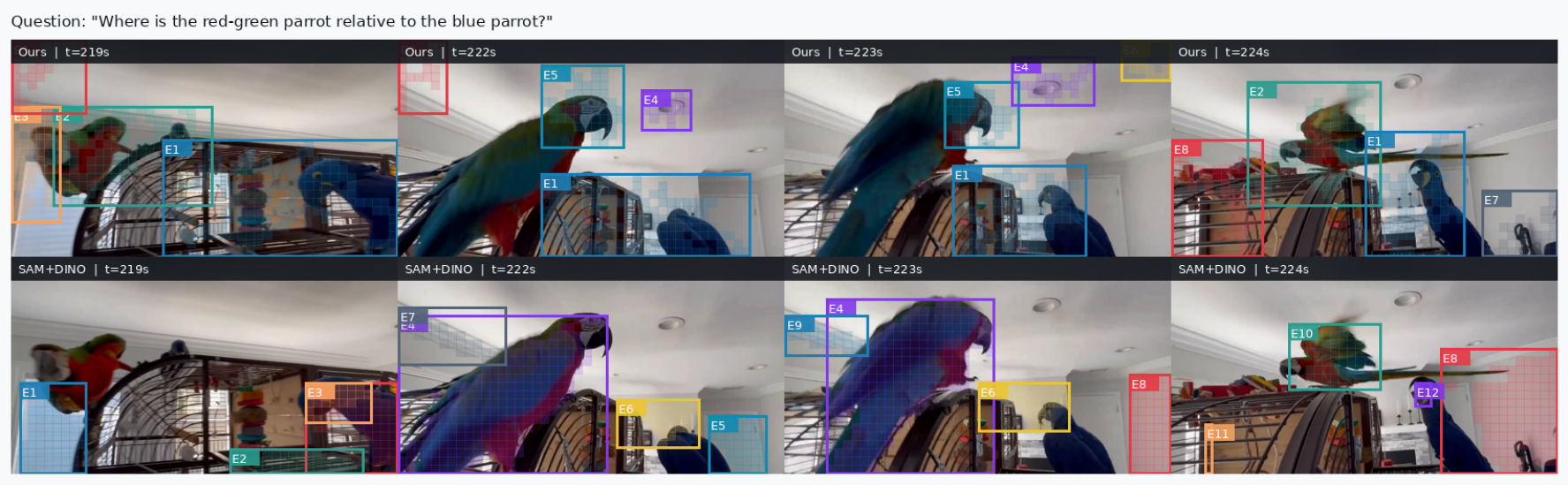}
    \caption{
        Qualitative analysis of latent object discovery.
        ObjectStream maintains task-relevant object states across consecutive
        observations, while SAM+DINOv2 produces finer masks but less consistent
        temporal associations.
    }
    \vspace{0.1in}
    \label{fig:visualization}
\end{figure*}

\paragraph{Object Memory Construction and Efficiency Analysis.}
We further analyze different object memory construction strategies under the same Qwen2.5-VL-7B backbone on OVO-Bench. As shown in Tab.~\ref{tab:object_construction_ablation}, Patch Top-K directly selects salient patch tokens without explicit object modeling. Although it keeps memory usage low, it yields lower performance and much higher TTFT than our method, showing that unstructured token selection is insufficient for streaming visual memory. DINOv2 Clustering introduces external patch features for object memory construction but brings additional memory and latency overhead while still underperforming our method. SAM+DINOv2 achieves the highest accuracy by using segmentation-based regions, yet it requires expensive segmentation and feature extraction, resulting in substantially higher latency and memory cost.

In contrast, ObjectStream discovers objects directly in the backbone visual token space and combines them with temporal retention. It achieves 73.3 performance with only 19.89 GB peak memory and 6.697s TTFT, offering a better balance between accuracy and efficiency. Fig.~\ref{fig:visualization} also illustrates this difference qualitatively. Although SAM+DINOv2 produces finer object masks, its outputs are optimized for pixel-level segmentation and do not directly provide a temporally reusable memory state for VLM reasoning. ObjectStream produces coarser but sufficiently localized object regions and maintains task-relevant objects across consecutive observations despite changes in scale, pose, and camera viewpoint. These results suggest that ObjectStream does not aim to replace segmentation models; instead, it provides a lightweight and temporally grounded object construction mechanism that better satisfies the low-latency and memory-bounded requirements of streaming video understanding.

\begin{table}[t]
\centering
\small
\setlength{\tabcolsep}{3.0pt}
\renewcommand{\arraystretch}{1.10}
\begin{tabular}{lcccc}
\toprule
\textbf{Variant}
& \textbf{\shortstack{Peak Mem$\downarrow$}}
& \textbf{TTFT$\downarrow$}
& \textbf{TPOT$\downarrow$}
& \textbf{Perf$\uparrow$} \\
\midrule
\multicolumn{5}{c}{\textit{Reference Baselines}} \\
\midrule
Qwen2.5-VL-7B
& 40.50GB
& 13.431s
& 0.336s
& 64.2 \\
+ FluxMem
& 25.78GB
& \textbf{6.641s}
& 0.252s
& 67.7 \\
\midrule
\multicolumn{5}{c}{\textit{object Construction Variants}} \\
\midrule
Patch Top-K
& \textbf{19.89GB}
& 16.223s
& 0.320s
& 72.8 \\
DINOv2 Cluster
& 23.09GB
& 18.120s
& 1.230s
& 72.3 \\
SAM+DINOv2
& 26.78GB
& 19.060s
& 0.785s
& \textbf{73.8} \\
\textbf{Latent Objects (Ours)} 
& \textbf{19.89GB}
& 6.697s
& \textbf{0.240s}
& 73.3 \\
\bottomrule
\end{tabular}
\caption{Analysis on different object memory construction strategies with Qwen2.5-VL-7B as backbone.}
\vspace{0.1in}
\label{tab:object_construction_ablation}
\end{table}

\section{Conclusion}
To conclude, in this paper, we propose ObjectStream, a training-free object-centric memory mechanism for efficient streaming video understanding. By discovering object-latent anchors and linking them into persistent tracks, ObjectStream builds a compact semantic object related memory over incoming video streams. This design preserves persistent visual evidence while remaining sensitive to dynamic changes and the latest visual state, enabling effective reasoning under memory and latency constraints. Experiments on both streaming and offline long-video benchmarks show that ObjectStream substantially improves online and offline video understanding while maintaining strong robustness under limited token budgets.

\newpage

\bibliography{aaai2027}
\bigskip
% \noindent Thank you for reading these instructions carefully. We look forward to receiving your electronic files!

% \bibliography{aaai2027}

% % Check whether the conference requires a reproducibility checklist to be included in the paper.
% % If so, you can uncomment the following line and ajust the path to include it.
% % \input{ReproducibilityChecklist.tex}

 \end{document}